\pdfoutput=1

\documentclass[table,xcdraw,11pt]{article}

\usepackage{EMNLP2022}

\usepackage{times}
\usepackage{latexsym}

\usepackage[T1]{fontenc}

\usepackage[utf8]{inputenc}

\usepackage{microtype}

\usepackage{booktabs}
\usepackage{multirow}
\usepackage{graphicx}
\usepackage{hyperref}

\usepackage{tabularx}
\usepackage{amssymb}
\usepackage{kantlipsum}

\usepackage{inconsolata}

\usepackage{dblfloatfix}

\title{Probing Commonsense Knowledge in Pre-trained Language Models \\ with Sense-level Precision and Expanded Vocabulary}

\author{Daniel Loureiro$^{\diamondsuit\clubsuit}$, Al\'ipio M\'ario Jorge$^{\clubsuit}$ \\
$^{\diamondsuit}$ Cardiff NLP, School of Computer Science and Informatics, Cardiff University, UK  \\
  $^{\clubsuit}$LIAAD - INESC TEC, Faculty of Sciences, University of Porto, Portugal \\
  {\tt boucanovaloureirod@cardiff.ac.uk, amjorge@fc.up.pt}}

\begin{document}
\maketitle
\begin{abstract}

Progress on commonsense reasoning is usually measured from performance improvements on Question Answering tasks designed to require commonsense knowledge.
However, fine-tuning large Language Models (LMs) on these specific tasks does not directly evaluate commonsense learned during pre-training. 
The most direct assessments of commonsense knowledge in pre-trained LMs are arguably cloze-style tasks targeting commonsense assertions (e.g., A pen is used for \texttt{[MASK]}.).
However, this approach is restricted by the LM's vocabulary available for masked predictions, and its precision is subject to the context provided by the assertion.
In this work, we present a method for enriching LMs with a  grounded sense inventory (i.e., WordNet) available at the vocabulary level, without further training.
This modification augments the prediction space of cloze-style prompts to the size of a large ontology while enabling finer-grained (sense-level) queries and predictions.
In order to evaluate LMs with higher precision, we propose SenseLAMA, a cloze-style task featuring verbalized relations from disambiguated triples sourced from WordNet, WikiData, and ConceptNet.
Applying our method to BERT, producing a WordNet-enriched version named SynBERT, we find that LMs can learn non-trivial commonsense knowledge from self-supervision, covering numerous relations, and more effectively than comparable similarity-based approaches.
\end{abstract}

\section{Introduction}


A relatively new direction for benchmarking Language Models (LMs) are tasks designed to require commonsense knowledge and reasoning.
These tasks usually target commonsense concepts under a Question Answering (QA) format \cite{mihaylov-etal-2018-suit,talmor-etal-2019-commonsenseqa,Bisk2020,nie-etal-2020-adversarial} and follow scaling trends.
Increasing the model's parameters leads to improved results, specially in few-shot learning settings \cite{palm}. 
Hybrid methods, particularly those fusing LMs with Graph Neural Networks, have shown that Commonsense Knowledge Graphs (CKGs) can help improve performance on these tasks \cite{xu-etal-2021-fusing,yasunaga-etal-2021-qa,zhang2022greaselm}.
The results obtained by these works, using relatively small LMs, suggest that CKGs can be an alternative (or complement) to increasing model size, with the added benefit of supporting more interpretable results.

Nevertheless, the QA approach provides only an indirect measure of a pre-trained model's ability to understand and reason with commonsense concepts.
The models attaining best results on these tasks are often too large for thorough analysis, and the QA format can promote shallow learning from annotation artifacts or spurious cues unrelated to commonsense \cite{branco-etal-2021-shortcutted}.

There are more direct ways of evaluating commonsense knowledge in LMs, such as scoring generated triples \cite{davison-etal-2019-commonsense}, infilling cloze-style statements \cite{petroni-etal-2019-language}, or fine-tuning for explicit generation of commonsense statements \cite{bosselut-etal-2019-comet}.
However, these approaches are either limited by each LM's particular vocabulary or biased by the available training data \cite{wang-etal-2021-language-models}.
Additionally, existing tasks and methods do not target grounded representations, which is essential for high-precision CKGs \cite{webchild,dalvi-mishra-etal-2017-domain}, and context-independent reference \cite{eyal-etal-2022-large}.

Commonsense tasks and approaches typically leverage ConceptNet \cite{conceptnet55}, a popular CKG built from an extensive crowdsourcing effort \cite{storks_survey}.
Although ConceptNet is arguably the most popular CKG available, its nodes are composed of free-form text rather than disambiguated (canonical) representations, allowing for misleading associations and aggravating the network's sparsity  \cite{li-etal-2016-commonsense,jastrzebski-etal-2018-commonsense,wang-etal-2020-connecting}.
The WordNet \cite{Miller1992WordNetAL} sense inventory is a natural choice for a set of ontologically grounded concept-level representations, having been curated by experts over decades and spanning various knowledge domains and syntactic categories of the English language.
Recent developments on WSD and Uninformed Sense Matching (USM) have shown that WordNet senses can be mapped to naturally occurring sentences with high precision \cite{LOUREIRO2022103661}, including at higher-abstraction levels (e.g., `Marlon Brando' to actor$_n^1$).
WordNet's utility for commonsense tasks is limited by its narrow set of relations, focused on lexical relations (mostly hypernymy).
However, its smaller size, compared to WikiData \cite{wikidata} or BabelNet \cite{NAVIGLI2012217}, for example, also presents an opportunity for effective expansion with reduced sparsity, which is important for symbolic reasoning \cite{NEURIPS2021_d367eef1}.

\begin{figure}[tb]
    \centering
    \includegraphics[width=7cm]{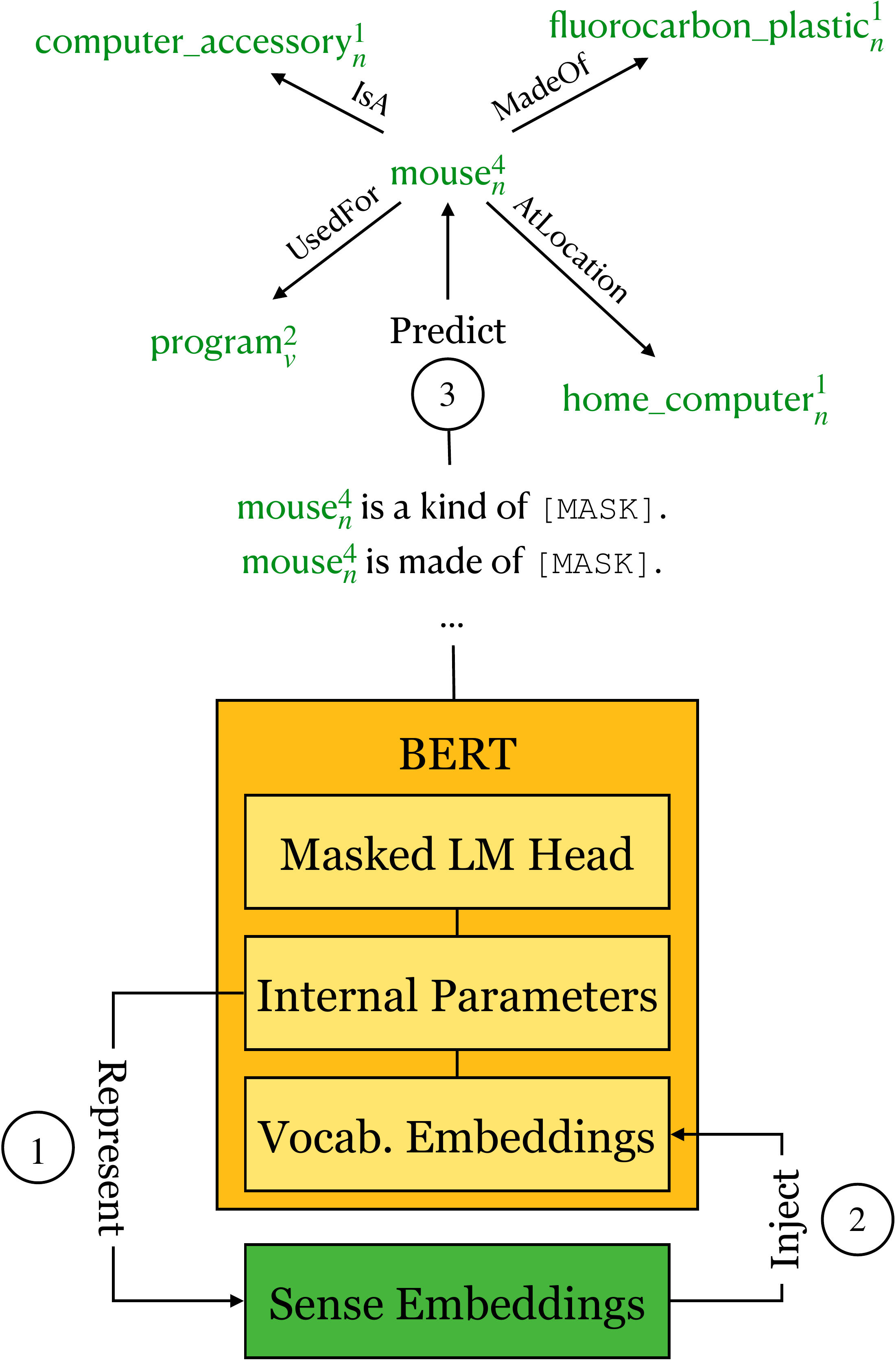}
    \caption{Our 3-step method for extracting unsupervised commonsense relations between concepts (i.e., word senses) from pre-trained language models. Relations are expressed as verbalizations that may be exchanged to target any other property of interest.}
    \label{fig:overview}
\end{figure}

In this work, we propose that a LM augmented with explicit sense-level representations (see \autoref{fig:overview}) may present a solution for precise evaluation of commonsense knowledge learned during pre-training that is not limited by the LM's vocabulary. Additionally, we explore how this enriched model can be used for grounded commonsense relation extraction towards precise and unbiased (w.r.t. commonsense training data) CKG construction that hybrid approaches may use.
Considering there is currently no set of grounded assertions available to assess progress in this direction, we propose a cloze-style probing task targeting specific senses and commonsense relations, inspired by \citet{petroni-etal-2019-language}.
Our contributions\footnote{\url{https://github.com/danlou/synbert}} are the following:
\begin{itemize}
  \item A BERT\footnote{While we focus on BERT and WordNet, our methods are broadly applicable to LMs and alternative representations.} model with 117k new sense-specific embeddings added to its vocabulary, based on the model's own internal states (SynBERT).
  \item The SenseLAMA probing task targeting wide-ranging and precise commonsense -- based on WordNet, WikiData, and ConceptNet.
  \item Analyses on the impact of different input types for eliciting accurate commonsense knowledge from BERT.
  \item A new CKG grounded on WordNet with 23k unseen triples over 18 commonsense relations (e.g., \textit{UsedFor}) generated by prompting.
\end{itemize}


\section{Related Work}

Large LMs have featured prominently in the latest efforts to build richer and more accurate CKGs.
COMET \cite{Hwang2021COMETATOMIC2O} is a generative model based on BART \cite{lewis-etal-2020-bart} trained on ConceptNet and ATOMIC \cite{DBLP:conf/aaai/SapBABLRRSC19} and proven capable of producing novel accurate triples for challenging relation types, such as \textit{HinderedBy}.
More recently, \citet{DBLP:journals/corr/abs-2110-07178} have proposed ATOMIC-10x, which leverages generated text from GPT-3 \cite{NEURIPS2020_1457c0d6} in combination with a critic model to create the largest and most accurate semi-automatically constructed CKG.
This accuracy was determined using both qualitative human ratings and quantitative measures.
However, these works are primarily concerned with extracting large CKGs using fine-tuned or distilled LMs, and do not focus on directly evaluating the CSK learned during pre-training.
Additionally, these works do not target grounded representations, considering only relations between free-text nodes, similarly to ConceptNet.

Addressing both disambiguated representations and sparsity resulting from free-text redundancy, WebChild \cite{webchild} proposes a CKG, grounded on WordNet senses, assembled from label propagation and pattern matching on Web corpora. WebChild features a large CKG (over 4M triples), but it predates large contextual LMs and the ensuing progress in WSD, making this resource unreliable by current standards.
Recent works on CKGs also focus on other aspects besides size and accuracy, such as salience \cite{chalier2020joint} or alternatives to triples \cite{Nguyen2021AdvancedSF}.

Our work is most related to LAMA \cite{petroni-etal-2019-language}, which compiles masked assertions based on triples from ConceptNet and other resources, and measures how many triples can be accurately recovered when masking the object term.
However, LAMA was designed for single-token masked prediction based on the intersection of the subword or byte-level token vocabularies used by the particular set of LMs considered in that work\footnote{This limitation stems from the fact that each word may be split into several tokens, whose number conditions predictions to words that match it, and is specific to each LM's tokenizer.}.
Consequently, LAMA is limited by design to a total of 21k prediction candidates. 

LAMA is an important early result of LM probing, but besides the previously mentioned technical limitations, its findings have also been challenged in later works.
\citet{kassner-schutze-2020-negated} demonstrated that LMs are susceptible to mispriming and often unable to handle negation.
\citet{poerner-etal-2020-e} further showed that LMs could be biased by the surface form of entity names.
Moreover, \citet{dufter-etal-2021-static} found that static embeddings using a nearest neighbors ($k$-NN) approach can outperform LMs on the LAMA benchmark, casting doubt on the presumed advantages of large LMs for the task.
Still, LAMA inspired others to use knowledge graphs (KGs) generated by LMs for intrinsic evaluation.
\citet{swamy2021interpreting} proposes extracting KGs from LMs to support interpretability and direct comparison between different LMs, or training stages.
\citet{aspillaga2021tracking} follows a similar direction but  proposes evaluating extracted KGs by concept relatedness, using hypernymy relations from WordNet and sense-tagged glosses.

Our approach overcomes the vocabulary limitations of LAMA while outperforming a comparable $k$-NN baseline.
We also explore using extracted CKGs to evaluate LMs, alongside the generation of novel CKGs.


\section{SenseLAMA}
\label{sec:senselama}

We begin by describing our probing task to evaluate the commonsense knowledge learned during LM pre-training.
SenseLAMA features verbalized relations\footnote{\autoref{sec:appendix_templates} shows handcrafted templates used for WordNet and WikiData triples, following \citet{petroni-etal-2019-language}.} between word senses from triples sourced from WordNet, WikiData, and ConceptNet.
In the following, we describe how we compiled SenseLAMA using these resources, including mapping triples to specific WordNet senses (i.e., synsets).

Unlike other works (e.g., \citealp{feng-etal-2020-scalable}), we do not  merge similar relations. Since our approach is unsupervised, we do not benefit from additional examples per relation. Thus, we prefer preserving performance metrics specific to each source.

We use the core WordNet synsets, initially defined by \citet{f47895ce0ff749618a360186b065556d}, to create an easier subset of SenseLAMA. While the full WordNet covers over 117k synsets, core synsets  are restricted to the 5k\footnote{Only 4,960 synsets can be mapped to WordNet v3.0.} most frequently occurring word senses, dramatically reducing the number of prediction candidates.
Thus, our `Core' subset is derived from the `Full' SenseLAMA, including only instances where both arguments of the triple belong to the set of core WordNet synsets.
If this filter results in a relation with less than ten instances, that relation is discarded from the `Core' subset.
\autoref{tab:counts} reports counts for each source and relation in SenseLAMA.

\paragraph{WordNet}
Our base ontology already contains several relations which arguably fall under the scope of commonsense knowledge, such as hypernymy, meronymy, or antonymy. Since these relations already target synsets within WordNet, no additional mapping or disambiguation is required.
Very frequent relations are capped at 10k samples.

\paragraph{WikiData}
This vast resource contains millions of triples for thousands of relations.
We only consider a few select relations most associated with commonsense knowledge.
Furthermore, we only admit triples for which the head and tail can be mapped to WordNet v3.0, either via the direct link available in WikiData's item properties or through linking to BabelNet, which we map to WordNet using the mapping from \citet{NAVIGLI2012217}.
Alternatively, we map some triples via hapax linking \cite{mccrae-cillessen-2021-towards}, when the triple's arguments correspond to unambiguous words.

\paragraph{ConceptNet}
We focus on the OMCS subset of ConceptNet, which includes full sentences collected through crowdsourcing, together with the corresponding triples.
Using these sentences, we do not require templates and can provide systems with the same input presented to crowd workers.
For each triple and sentence pair, we align the triple's arguments to the sentence and disambiguate those words in context using ESC \cite{barba-etal-2021-esc}, a state-of-the-art WSD system.
Triples that cannot be successfully aligned are discarded.
For added precision, we constrain WSD according to each relation's particular Part-of-Speech types, following \citet{havasi2009digital}.

\begin{table}[th]
    \centering
    \resizebox{0.70\columnwidth}{!}{%
    \begin{tabular}{@{}lccl@{}}
    \toprule
    \textbf{Source/Relation} & \multicolumn{1}{c}{\textbf{Core}} & \multicolumn{1}{c}{\textbf{Full}} \\ \midrule    
    \textbf{WordNet (WN)} & 1,757 & 41,237 \\
    \textit{Hypernym} & 1,288 & 10,000 \\
    \textit{Holonym (Member)} & 26 & 10,000 \\
    \textit{Holonym (Part)} & 145 & 7,832 \\
    \textit{Antonym} & 282 & 7,391 \\
    \textit{Hypernym (Instance)} & - & 5,356 \\
    \textit{Meronym (Substance)} & 16 & 658 \\ \midrule
    
    \textbf{WikiData (WD)} & 136 & 7,222 \\
    \textit{P31 (Instance of)} & 39 & 2,968 \\
    \textit{P361 (Part of)} & 45 & 1,367 \\
    \textit{P366 (Use)} & 27 & 763 \\
    \textit{P186 (Made from)} & 15 & 639 \\
    \textit{P461 (Opposite of)} & 10 & 501 \\
    \textit{P737 (Influenced by)} & - & 316 \\
    \textit{P2283 (Uses)} & - & 268 \\
    \textit{P463 (Member of)} & - & 183 \\
    \textit{P1535 (Used by)} & - & 151 \\
    \textit{P279 (Subclass of)} & - & 66 \\ \midrule
    
    \textbf{ConceptNet (CN)} & 1,032 & 3,541 \\
    \textit{AtLocation} & 328 & 1,004 \\
    \textit{UsedFor} & 162 & 661 \\
    \textit{IsA} & 120 & 512 \\
    \textit{Causes} & 38 & 224 \\
    \textit{HasSubevent} & 57 & 205 \\
    \textit{HasPrerequisite} & 50 & 165 \\
    \textit{HasProperty} & 47 & 157 \\
    \textit{CapableOf} & 48 & 123 \\
    \textit{MotivatedByGoal} & 37 & 104 \\
    \textit{HasA} & 48 & 97 \\
    \textit{PartOf} & 33 & 80 \\
    \textit{CausesDesire} & 14 & 52 \\
    \textit{ReceivesAction} & 19 & 44 \\
    \textit{MadeOf} & 18 & 42 \\
    \textit{Desires} & 13 & 28 \\
    \textit{CreatedBy} & - & 17 \\
    \textit{HasFirstSubevent} & - & 14 \\
    \textit{HasLastSubevent} & - & 12 \\ \midrule
    \textbf{All} & 2,925 & 52,000 \\ 
    \bottomrule
    \end{tabular}%
    }
\caption{SenseLAMA relation counts.}
\label{tab:counts}
\end{table}

\section{SynBERT}
\label{sec:synbert}

In this section, we cover the three steps employed to enrich LMs with sense embeddings: 1) Represent word senses from internal states; 2) Map and add sense embeddings to the LM's vocabulary; 3) Adapt cloze-style assertions and predictions to extract grounded triples.
See \autoref{fig:overview} for an overview.
Throughout this work, we use BERT-Large \cite{devlin-etal-2019-bert} as our reference LM.

\subsection{Sense Representation}
\label{sec:synbert_repr}

For representing word senses with LMs, we follow \citet{LOUREIRO2022103661} and learn sense embeddings as centroids of contextual embeddings from sense-annotated corpora and glosses.
We follow the recommendation of representing contextual embeddings with weighted pooling from all layers, using weights specific to the sense matching profile (i.e., LMMS SP-USM).
We also average the embeddings from annotations with gloss embeddings (centroids of contextual embeddings for lemmas and tokens in each synset's gloss).
These sense embeddings are derived from a LM's frozen parameters, relying exclusively on modeling capability learned during pre-training.
These sense embeddings have demonstrated state-of-the-art performance across several sense-related tasks, without bias towards most frequent senses, as observed with fine-tuning approaches \cite{loureiro-etal-2021-analysis}.

\subsection{Mapping and Injecting Embeddings}
\label{sec:synbert_map}

\citet{poerner-etal-2020-e} found that linear mapping was sufficient for high accuracy alignment between static embeddings (unrelated to BERT) and BERT's vocabulary embeddings.
We follow this approach since our sense embeddings are derived from BERT, making alignment theoretically more straightforward.
In order to learn the linear mapping (using least-squares), we need tokens represented in both the LM's vocabulary embedding space (i.e., input-space) and the alternate space defined by the weighted pooling of layers used to represent the sense embeddings.
We obtain this by finding tokens in the LM's vocabulary with more than 100 occurrences in Wikitext \cite{DBLP:journals/corr/MerityXBS16} and applying the same pooling used for sense representation to learn embeddings for those tokens in the alternate space.
After mapping, sense embeddings are added to the LM's vocabulary as special tokens, represented using a distinct format (<WN:\textit{synset}>), similarly to \citet{schick-schutze-2020-bertram}.

\begin{table*}[!t]
\centering
\resizebox{\textwidth}{!}{%
\begin{tabular}{llll}
\toprule
& \multicolumn{1}{c}{\textbf{IsA}} & \multicolumn{1}{c}{\textbf{Desires}} & \multicolumn{1}{c}{\textbf{MadeOf}} \\ \midrule
\textbf{} & \large{A \textbf{mouse} is a kind of \texttt{[MASK]} .} & \large{A \textbf{mouse} wants to \texttt{[MASK]} .} & \large{A \textbf{mouse} is made of \texttt{[MASK]} .} \\
\small{\textbf{BERT}} & animal, rabbit, cat & play, eat, talk & wood, clay, bone \\
\small{\textbf{SynBERT}} & mouse-eared\_bat$_{n}^{1}$, mouser$_{n}^{1}$, rabbit\_ears$_{n}^{2}$ & die$_{v}^{2}$, forage$_{v}^{2}$, feed$_{v}^{7}$ & redwood$_{n}^{1}$, wood$_{n}^{1}$, yellowwood$_{n}^{1}$ \\ \midrule
 & \large{A \textbf{mouse$_{n}^{1}$} is a kind of \texttt{[MASK]} .} & \large{A \textbf{mouse$_{n}^{1}$} wants to \texttt{[MASK]} .} & \large{A \textbf{mouse$_{n}^{1}$} is made of \texttt{[MASK]} .} \\
\small{\textbf{SynBERT}} & rat$_{n}^{1}$, mouse-eared\_bat$_{n}^{1}$, pocket\_rat$_{n}^{1}$ & forage$_{v}^{2}$, feed$_{v}^{7}$, die$_{v}^{2}$ & round\_bone$_{n}^{1}$, bone$_{n}^{1}$, leg$_{n}^{2}$ \\ \midrule
 & \large{A \textbf{mouse$_{n}^{4}$} is a kind of \texttt{[MASK]} .} & \large{A \textbf{mouse$_{n}^{4}$} wants to \texttt{[MASK]} .} & \large{A \textbf{mouse$_{n}^{4}$} is made of \texttt{[MASK]} .} \\
\small{\textbf{SynBERT}} & computer\_keyboard$_{n}^{1}$, computer\_accessory$_{n}^{1}$,  & move$_{v}^{13}$, move$_{v}^{12}$, think$_{v}^{6}$ & fluorocarbon\_plastic$_{n}^{1}$, glass$_{n}^{1}$, \\
 & computer\_memory\_unit$_{n}^{1}$ &  & wire\_glass$_{n}^{1}$ \\ \midrule
 & \large{A \textbf{mouse$_{n}^{3}$} is a kind of \texttt{[MASK]} .} & \large{A \textbf{mouse$_{n}^{3}$} wants to \texttt{[MASK]} .} & \large{A \textbf{mouse$_{n}^{3}$} is made of \texttt{[MASK]} .} \\
\small{\textbf{SynBERT}} & dummy$_{n}^{1}$, shy\_person$_{n}^{1}$, small\_person$_{n}^{1}$ & shop\_talk$_{n}^{1}$, talk$_{n}^{3}$, talk$_{n}^{2}$ & redwood$_{n}^{1}$, ironwood$_{n}^{2}$, yellowwood$_{n}^{1}$ \\
\bottomrule
\end{tabular}%
}
\caption{Top-3 masked predictions targeting `mouse' using templates corresponding to the \textit{IsA}, \textit{Desires} and \textit{MadeOf} relations. First row does not use special synset tokens in the input and shows predictions using BERT (ignoring stopwords) as well SynBERT (ignoring regular tokens). Next rows show predictions using special tokens corresponding to the 3 senses for `mouse' available in WordNet. Their definitions are the following: mouse$_{n}^{1}$ - any of numerous small rodents typically resembling diminutive rats [...]; mouse$_{n}^{4}$ - a hand-operated electronic device that controls the coordinates of a cursor on your computer screen [...]; mouse$_{n}^{3}$ - person who is quiet or timid.}
\label{tab:preds}
\end{table*}

\subsection{Extracting Triples}
\label{sec:synbert_extract}

Triples are extracted using assertions relating a grounded word sense with a masked tail term (e.g., [pen$_{n}^{1}$, \textit{UsedFor}, ?] $\rightarrow$ "A <WN:\textit{pen.n.01}> can be used for \texttt{[MASK]}.").
Regular tokens are discarded from the LM's masked predictions\footnote{The head term is also removed from predictions.}, and softmax normalization is performed after filtering so that prediction scores are distributed exclusively over grounded word senses.
Our default setup\footnote{See \autoref{sec:analysis_repr} for an ablation analysis.} prepends assertions with the head term's gloss from WordNet for improved results (i.e., "<WN:\textit{synset}> can be defined as : \textit{gloss} . \texttt{[SEP]} \textit{assertion}").
We refer to \autoref{tab:preds} for example predictions.

\section{Experiments}

In this section, we explore two applications for our method that motivated this work: 1) Evaluating commonsense knowledge learned during LM pre-training; 2) Extracting precise CKGs from LMs enriched with grounded word senses.

\subsection{Probing with SenseLAMA}
\label{sec:exp_senselama}

The SenseLAMA probe described in \autoref{sec:senselama} is used to evaluate the commonsense knowledge learned while pre-training LMs, through the adaptation described in \autoref{sec:synbert}.
The prediction methodology described in \autoref{sec:synbert_extract} is used to obtain ranked predictions for tail terms masked in SenseLAMA.
Performance is evaluated using ranking metrics, namely mean Precision @ k and Mean Reciprocal Rank (MRR).
As with LAMA, many instances admit various possible answers (1 to N). Therefore P@10 may be considered more representative of actual performance than P@1.

The complete results in \autoref{tab:complete_results} show that performance varies substantially by source and relation.
It is interesting to note that for core synsets (i.e., frequent concepts), we find P@10 above 30\% for most relations.
Particular relation groups, such as \textit{Holonym (Part)}, \textit{P361 (Part of)} and \textit{PartOf} show particularly high results (above 60\% P@10), suggesting that extraction for these relation types could be reliable enough for some applications.

The Full set appears much more challenging, which is to be expected considering the 20x increase for the search space in this setting, along with several instances targeting rare concepts. While this setting is much less reliable, we still find that most relations can be accurately predicted from the top 1\% of candidates (> 60\% P@1000).

Out of 39 relations (Full set), the most challenging belong to ConceptNet, particularly \textit{ReceivesAction}, \textit{Desires}, \textit{CausesDesire} and \textit{HasSubevent}, supporting the claim that commonsense relations are harder to model by LMs than lexical relations.

\begin{table*}[tbp]
\centering
\resizebox{0.99\textwidth}{!}{%
\begin{tabular}{@{}l|ccccc||cccccc@{}}
\toprule
 & \multicolumn{5}{|c||}{\textbf{Core (4,960 candidates)}} & \multicolumn{6}{c}{\textbf{Full (117,659 candidates)}} \\
\multicolumn{1}{l|}{} & \multicolumn{1}{c}{\textbf{P@1}} & \multicolumn{1}{c}{\textbf{P@3}} & \multicolumn{1}{c}{\textbf{P@10}} & \multicolumn{1}{c}{\textbf{P@100}} & \multicolumn{1}{c||}{\textbf{MRR}} & \textbf{P@1} & \textbf{P@3} & \textbf{P@10} & \textbf{P@100} & \textbf{P@1000} & \textbf{MRR} \\ \midrule
\textbf{All} & 24.41 & 40.56 & 59.10 & 83.20 & 35.64 & 7.18 & 13.78 & 23.09 & 45.75 & 71.75 & 12.55 \\
 &  &  &  &  &  &  &  &  &  &  &  \\

\textbf{WordNet} & 31.25 & 49.80 & 69.10 & 87.82 & 43.46 & 7.78 & 14.75 & 24.26 & 46.39 & 71.84 & 13.34 \\
\textit{Hypernym} & 29.04 & 45.96 & 66.15 & 86.10 & 40.77 & 8.31 & 17.24 & 30.77 & 59.17 & 82.74 & 15.65 \\
\textit{Holonym (Member)} & 42.31 & 69.23 & 88.46 & 100.00 & 57.80 & 1.75 & 3.04 & 5.03 & 13.98 & 41.89 & 3.00 \\
\textit{Holonym (Part)} & 34.48 & 60.69 & 80.69 & 92.41 & 50.20 & 13.97 & 25.93 & 40.63 & 67.89 & 88.15 & 22.91 \\
\textit{Antonym} & 37.94 & 58.16 & 74.11 & 91.49 & 50.09 & 8.10 & 13.72 & 20.96 & 40.56 & 70.32 & 12.55 \\
\textit{Hypernym (Instance)} & - & - & - & - & - & 9.19 & 18.09 & 30.08 & 61.24 & 86.45 & 16.35 \\
\textit{Meronym (Substance)} & 43.75 & 81.25 & 81.25 & 100.00 & 59.14 & 2.43 & 6.23 & 12.46 & 33.13 & 65.50 & 6.00 \\

 &  &  &  &  &  &  &  &  &  &  &  \\
\textbf{WikiData} & 16.18 & 33.09 & 49.26 & 79.41 & 27.62 & 5.05 & 10.12 & 18.83 & 43.91 & 72.07 & 9.69 \\
\textit{P31 (Instance of)} & 10.26 & 23.08 & 23.08 & 61.54 & 16.94 & 2.90 & 6.74 & 13.61 & 37.77 & 68.56 & 6.67 \\
\textit{P361 (Part of)} & 15.56 & 35.56 & 62.22 & 82.22 & 30.26 & 8.71 & 16.17 & 27.21 & 55.89 & 79.37 & 14.86 \\
\textit{P366 (Use)} & 14.81 & 25.93 & 48.15 & 88.89 & 24.80 & 4.06 & 9.70 & 19.27 & 42.07 & 64.74 & 9.00 \\
\textit{P186 (Made from)} & 33.33 & 46.67 & 60.00 & 86.67 & 41.66 & 8.61 & 12.83 & 23.63 & 46.64 & 72.77 & 13.03 \\
\textit{P461 (Opposite of)} & 20.00 & 60.00 & 80.00 & 100.00 & 43.91 & 8.98 & 18.36 & 30.34 & 60.28 & 81.24 & 16.35 \\
\textit{P737 (Influenced by)} & - & - & - & - & - & 2.53 & 6.33 & 10.76 & 31.33 & 71.20 & 5.85 \\
\textit{P2283 (Uses)} & - & - & - & - & - & 4.85 & 8.58 & 14.93 & 37.31 & 66.42 & 8.42 \\
\textit{P463 (Member of)} & - & - & - & - & - & 1.64 & 2.73 & 15.30 & 40.44 & 86.34 & 5.51 \\
\textit{P1535 (Used by)} & - & - & - & - & - & 0.66 & 3.97 & 11.92 & 41.06 & 72.85 & 4.53 \\
\textit{P279 (Subclass of)} & - & - & - & - & - & 6.06 & 12.12 & 21.21 & 45.45 & 72.73 & 10.64 \\
 &  &  &  &  &  &  &  &  &  &  &  \\

\textbf{ConceptNet} & 13.86 & 25.87 & 43.51 & 75.78 & 23.38 & 4.55 & 9.88 & 18.07 & 42.11 & 70.06 & 9.23 \\
\textit{AtLocation} & 14.02 & 25.91 & 46.95 & 79.27 & 24.24 & 4.98 & 10.56 & 19.82 & 45.82 & 76.10 & 10.09 \\
\textit{UsedFor} & 7.41 & 16.67 & 36.42 & 75.93 & 16.04 & 3.18 & 8.17 & 15.13 & 38.88 & 69.59 & 7.48 \\
\textit{IsA} & 27.50 & 43.33 & 62.50 & 87.50 & 38.56 & 7.42 & 13.67 & 27.34 & 59.38 & 83.59 & 13.61 \\
\textit{Causes} & 5.26 & 23.68 & 34.21 & 65.79 & 16.56 & 2.68 & 6.25 & 12.05 & 27.68 & 54.91 & 5.84 \\
\textit{HasSubevent} & 3.51 & 14.04 & 19.30 & 43.86 & 10.08 & 0.98 & 2.44 & 5.37 & 14.63 & 35.12 & 2.64 \\
\textit{HasPrerequisite} & 4.00 & 16.00 & 26.00 & 78.00 & 13.16 & 3.64 & 8.48 & 13.94 & 41.21 & 71.52 & 7.35 \\
\textit{HasProperty} & 4.26 & 14.89 & 38.30 & 76.60 & 14.65 & 2.55 & 5.10 & 9.55 & 29.30 & 63.69 & 5.21 \\
\textit{CapableOf} & 8.33 & 18.75 & 33.33 & 54.17 & 16.09 & 2.44 & 7.32 & 13.82 & 30.08 & 51.22 & 6.48 \\
\textit{MotivatedByGoal} & 29.73 & 51.35 & 67.57 & 89.19 & 43.59 & 6.73 & 20.19 & 29.81 & 63.46 & 80.77 & 15.66 \\
\textit{HasA} & 16.67 & 27.08 & 41.67 & 81.25 & 24.35 & 11.34 & 16.49 & 23.71 & 48.45 & 79.38 & 16.07 \\
\textit{PartOf} & 36.36 & 54.55 & 75.76 & 90.91 & 48.35 & 10.00 & 21.25 & 37.50 & 66.25 & 87.50 & 19.33 \\
\textit{CausesDesire} & 0.00 & 7.14 & 28.57 & 78.57 & 7.61 & 0.00 & 1.92 & 5.77 & 30.77 & 65.38 & 2.44 \\
\textit{ReceivesAction} & 0.00 & 0.00 & 10.53 & 31.58 & 3.43 & 0.00 & 0.00 & 0.00 & 6.82 & 20.45 & 0.19 \\
\textit{MadeOf} & 44.44 & 50.00 & 61.11 & 100.00 & 51.52 & 9.52 & 30.95 & 33.33 & 47.62 & 80.95 & 19.34 \\
\textit{Desires} & 7.69 & 15.38 & 23.08 & 46.15 & 12.15 & 0.00 & 3.57 & 3.57 & 25.00 & 46.43 & 1.92 \\
\textit{CreatedBy} & - & - & - & - & - & 0.00 & 0.00 & 11.76 & 35.29 & 82.35 & 3.43 \\
\textit{HasFirstSubevent} & - & - & - & - & - & 7.14 & 7.14 & 14.29 & 35.71 & 78.57 & 9.23 \\
\textit{HasLastSubevent} & - & - & - & - & - & 0.00 & 0.00 & 16.67 & 33.33 & 58.33 & 2.38 \\

\bottomrule
\end{tabular}%
}
\caption{Complete results on the SenseLAMA probing task using BERT Large with LMMS SP-USM sense embeddings. Reporting Precision at k (P@k) and Mean Reciprocal Rank (MRR). Sorted by P@1 on the Full set.}
\label{tab:complete_results}
\end{table*}

\subsection{Commonsense Knowledge Extraction}
\label{sec:exp_extraction}

While it is possible to use the method presented in this work to exhaustively query LMs and rank predictions for every synset and relation, we take a simpler approach in this experiment.
Considering that the ConceptNet subset of SenseLAMA includes higher quality assertions (not generated by templates), we use these to generate new query assertions by replacing the head terms with their co-hyponyms. 
This approach also reduces the chances of generating non-sensical queries (e.g., "A <WN:\textit{pen.n.01}> desires \texttt{[MASK]}."), which would result from combinatorial generation.
Keeping in mind that there is likely more than a single valid prediction for each assertion, we use a threshold to extract multiple triples from each assertion's prediction distribution.
This threshold is automatically determined as the median score assigned to correct predictions on the SenseLAMA (Full) probe.

This process generates 36,505 query assertions and 23,088 novel\footnote{Not part of the triples in SenseLAMA or its sources.} triples  scoring above the threshold.
This novel CKG, grounded on WordNet, covers 18 commonsense relations and reaches 9.2\% of all synsets.
See \autoref{sec:appendix_extraction} for detailed statistics.

\section{Analysis}
\label{sec:analysis}

The analyses reported in this section focus on the following comparisons: 1) Alternatives for representing triples as cloze-style assertions; 2) Verbalization against Nearest Neighbors; 3) Mapping embeddings or retaining geometry.

\subsection{Triple Representation}
\label{sec:analysis_repr}

For this analysis, we compare alternatives for representing triples as masked assertions, specifically using synsets (special tokens) instead of regular tokens (i.e., most frequent lemma\footnote{Each synset may be associated to multiple lemmas. Frequencies obtained from wordfreq \cite{robyn_speer_2018_1443582}.}) and glosses (averaged with sense embeddings and prepended to the assertion).
We also combine lemmas and synsets using the \textit{slash} representation \cite{schick-schutze-2020-bertram}, where the head term in the assertion is replaced with "\textit{lemma} / <WN:\textit{synset}>".

Results in \autoref{tab:ablation} show that the synset representation is more effective than lemmas, and while the combination of lemmas and synsets is better, using synsets exclusively provides slightly improved results.
Glosses appear to have a substantial impact under all settings, but averaging gloss embeddings and prepending glosses shows the best results.
These results also show that ConceptNet (CN) is not only the most challenging subset but also the least sensitive to these experimental choices.
For completeness, \autoref{sec:appendix_senselama_no_gloss} reports complete SenseLAMA results using sense embeddings (from annotated text) that have not been merged with gloss embeddings or used assertions prepended with the gloss for the head synset.

\begin{table}[ht]
\centering
\resizebox{0.49\textwidth}{!}{%
\begin{tabular}{@{}cc|cc|cccc@{}}
\toprule
\multicolumn{2}{c}{\textbf{Token}} & \multicolumn{2}{c}{\textbf{Gloss}} &  &  &  &  \\
\small{\textbf{Lem}} & \small{\textbf{Syn}} & \small{\textbf{Avg}} & \small{\textbf{Pre}} & \multirow{-2}{*}{\textbf{WN}} & \multirow{-2}{*}{\textbf{WD}} & \multirow{-2}{*}{\textbf{CN}} & \multirow{-2}{*}{\textbf{ALL}} \\ \midrule
\checkmark &  &  &  & 19.74 & 17.32 & 16.52 & 18.49 \\
\checkmark &  & \checkmark &  & 20.79 & 16.62 & 17.40 & 19.40 \\
\checkmark &  &  & \checkmark & 36.70 & 27.12 & 22.33 & 31.19 \\
\checkmark &  & \checkmark & \checkmark & 40.82 & \textbf{29.79} & 23.18 & 34.09 \\ \midrule
 & \checkmark &  &  & 26.46 & 19.46 & 17.67 & 23.04 \\
 & \checkmark & \checkmark &  & 30.39 & 20.72 & 18.60 & 25.78 \\
 & \checkmark &  & \checkmark & 38.76 & 26.83 & 22.56 & 32.49 \\
 & \checkmark & \checkmark & \checkmark & \textbf{43.44} & 27.59 & \textbf{23.38} & \textbf{35.63} \\ \midrule
\checkmark & \checkmark &  &  & 25.47 & 19.65 & 18.08 & 22.59 \\
\checkmark & \checkmark & \checkmark &  & 27.44 & 20.70 & 19.13 & 24.20 \\
\checkmark & \checkmark &  & \checkmark & 39.14 & 26.49 & 21.66 & 32.39 \\
\checkmark & \checkmark & \checkmark & \checkmark & 42.31 & 28.15 & 21.97 & 34.47 \\
\bottomrule
\end{tabular}%
}
\caption{MRR on SenseLAMA (Core) when representing lemmas (Lem) and/or synsets (Syn); averaging gloss embedding (Avg) and/or prepending the gloss (Pre).}
\label{tab:ablation}
\end{table}

\subsection{Impact of Verbalization}
\label{sec:analysis_verbalization}

\citet{dufter-etal-2021-static} showed that a nearest neighbors ($k$-NN) baseline using static embeddings could outperform BERT on the LAMA probe under comparable settings.
We run a similar experiment to verify whether the same conclusion may apply to our SenseLAMA probe and SynBERT model.

In our case, the sense embeddings (mapped) added to BERT's vocabulary can be used as static embeddings.
Bearing in mind that some sense embeddings from LMMS are inferred from hypernymy relations in WordNet (17.1\% of senses, mostly rare), we also experiment with another set of BERT-based sense embeddings which are not derived from any relations (ARES, \citealp{scarlini-etal-2020-contexts}).
For a fair comparison, we do not prepend glosses for masked predictions.

As such, \autoref{tab:core_summary} reports results using $k$-NN with sense embeddings, alongside using SynBERT with the verbalized queries (i.e., masked assertions) provided with SenseLAMA.
We verify that $k$-NN can outperform SynBERT under these conditions, but only for the more lexical-oriented relations in WordNet.
For ConceptNet, the source most strictly related to commonsense knowledge, we find verbalized queries provide a clear advantage over $k$-NN.
To a lesser extent, the encyclopedic triples of WikiData are also more accurately predicted with SynBERT.
This finding is in line with previous work comparing relational knowledge in BERT \cite{Bouraoui2020InducingRK}.

\begin{table*}[ht]
\centering
\resizebox{\textwidth}{!}{%
\begin{tabular}{@{}l|ccc|ccc|ccc|ccc@{}}
\toprule
 & \multicolumn{3}{c|}{\textbf{WordNet}} & \multicolumn{3}{c|}{\textbf{WikiData}} & \multicolumn{3}{c|}{\textbf{ConceptNet}} & \multicolumn{3}{c}{\textbf{All}} \\
 & \textbf{P@1} & \textbf{P@10} & \textbf{MRR} & \textbf{P@1} & \textbf{P@10} & \textbf{MRR} & \textbf{P@1} & \textbf{P@10} & \textbf{MRR} & \textbf{P@1} & \textbf{P@10} & \textbf{MRR} \\ \midrule
\textbf{Distance-based ($k$-NN)} &  &  &  &  &  &  &  &  &  &  &  &  \\
ARES & 17.87 & 58.91 & 31.07 & 9.56 & 36.76 & 18.27 & 3.97 & 20.45 & 9.31 & 12.58 & 44.31 & 22.80 \\
LMMS SP-USM & \textbf{26.12} & \textbf{63.18} & \textbf{38.19} & 8.09 & 34.56 & 16.67 & 3.10 & 17.44 & 7.97 & \textbf{17.16} & \textbf{45.71} & \textbf{26.53} \\
 &  &  &  &  &  &  &  &  &  &  &  &  \\
\textbf{Template-based (LM)} &  &  &  &  &  &  &  &  &  &  &  &  \\
ARES & 19.86 & 46.96 & 29.08 & \textbf{13.24} & \textbf{38.97} & \textbf{21.94} & 9.69 & 36.05 & 17.63 & 15.97 & 42.74 & 24.71 \\
LMMS SP-USM & 21.17 & 49.57 & 30.39 & 12.50 & 37.50 & 20.72 & \textbf{10.17} & \textbf{36.14} & \textbf{18.60} & 16.89 & 44.27 & 25.78 \\
\bottomrule
\end{tabular}%
}
\caption{Performance comparison on SenseLAMA (Core) using the baseline $k$-NN distance-based method (ignores relation) and the masked LM template-based method (verbalizes relation). For fair comparison, gloss prepending (see \autoref{sec:synbert_extract}) is not used for LM results.}
\label{tab:core_summary}
\end{table*}

\subsection{Degradation from Mapping}
\label{sec:analysis_mapping}

Our SynBERT model features sense embeddings that result from the straightforward linear mapping of embeddings pooled from all layers into the vocabulary embedding space (see \autoref{sec:synbert_map}).

For this analysis, we estimate the performance impact of this mapping procedure by comparing the performance of mapped and unmapped LMMS sense embeddings on the $k$-NN baseline for SenseLAMA (described on \autoref{sec:analysis_verbalization}).

Results on \autoref{tab:mapping} show that while the procedure is simple, mapped embeddings retain very similar performance to their original versions, with around 5\% degradation on P@1, P@10, and MRR.

\begin{table}[ht]
\centering
\begin{tabular}{cccc}
\toprule
 & \textbf{P@1} & \textbf{P@10} & \textbf{MRR} \\ \midrule
\textbf{Original} & 17.16 & 45.71 & 26.53 \\ \midrule
\multirow{2}{*}{\textbf{Mapped}} & 16.21 & 44.21 & 25.31 \\
 & (-5.5\%) & (-3.3\%) & (-4.6\%) \\
\bottomrule
\end{tabular}
\caption{Performance comparison on SenseLAMA (Core) using $k$-NN with original and mapped LMMS SP-USM embeddings.}
\label{tab:mapping}
\end{table}

\section{Conclusion}

We have shown that commonsense knowledge learned during LM pre-training can be probed more precisely and extensively using sense embedding learned from grounded ontologies, compared to prior work such as LAMA, which is limited to a subset of the LM's word-level vocabulary.


The proposed SynBERT model, adapted from BERT, along with SenseLAMA, our new probing task grounded on WordNet, provide clearer insights into which commonsense relations are best understood by LMs, and how the commonsense domain compares against more lexical or encyclopedic knowledge.
We also explore how SynBERT, or similar models, can be used to extract novel CKGs which may support recent hybrid methods fusing CKGs and LMs (e.g., \citealp{zhang2022greaselm}), or enable symbolic-first methods (e.g., \citealp{NEURIPS2021_d367eef1}) to leverage precise commonsense knowledge learned from self-supervision by LMs.

This paper is focused on establishing our approach using BERT as our reference model, due to its popularity related to probing LMs.
We leave a thorough comparison between BERT and alternative LMs for future work.

\section*{Limitations}

Although commonsense knowledge should remain mostly unchanged over time, the sense representations introduced in SynBERT, and targeted by SenseLAMA, are limited to the release date of WordNet v3.0 (2006).
As noted by \citet{eyal-etal-2022-large}, novel concepts that have recently become mainstream (e.g., covid) are missing from WordNet, therefore our contributions also do not cover these more recent concepts and relations.
Finally, this work should also not be considered from the standpoint of improving LM performance on downstream tasks (e.g., SuperGLUE).
Since the sense embeddings used are based on internal states of the LM, their integration at the vocabulary-level is not expected to add new information to the LM.

\section*{Reproducibility}

SynBERT, SenseLAMA, and related code, are freely available at \url{https://github.com/danlou/synbert} (MIT License).
Reported experiments run in under 3 hours on a RTX 3090, using a maximum of 12GB VRAM.
SynBERT, with the full set of 117k synsets, contains 454M parameters.

\bibliography{anthology,custom}
\bibliographystyle{acl_natbib}

\appendix

\clearpage

\section{Templates}
\label{sec:appendix_templates}

\begin{table*}[htb]
\centering
\resizebox{\textwidth}{!}{
\begin{tabular}{cllll}
\toprule
\textbf{Source} & \textbf{Relation} & \textbf{Template} & \multicolumn{2}{l}{\textbf{Example Pair ({[}H{]}ead, {[}T{]}ail)}} \\ \midrule
\multirow{6}{*}{WordNet} & Hypernym & {[}H{]} is a type of {[}T{]} & medicine$_n^2$ & drug$_n^1$ \\
 & Holonym (Member) & {[}H{]} is a member of {[}T{]} & princess$_n^1$ & royalty$_n^2$ \\
 & Holonym (Part) & {[}H{]} is part of {[}T{]} & jaw$_n^1$ & skull$_n^1$ \\
 & Antonym & {[}H{]} is the opposite of {[}T{]} & straight$_n^8$ & curved$_a^1$ \\
 & Hypernym (Instance) & {[}H{]} is an example of {[}T{]} & sahara$_n^1$ & desert$_n^1$ \\
 & Meronym (Substance) & {[}H{]} is made of {[}T{]} & bread$_n^1$ & flour$_n^1$ \\ \midrule
\multirow{10}{*}{WikiData} & P31 (Instance of) & {[}H{]} is an example of {[}T{]} & capitalism$_n^1$ & political\_orientation$_n^1$ \\
 & P361 (Part of) & {[}H{]} is part of {[}T{]} & regulation$_n^1$ & politics$_n^1$ \\
 & P366 (Use) & {[}H{]} is used for {[}T{]} & vegetable\_oil$_n^1$ & makeup$_n^1$ \\
 & P186 (Made from) & {[}H{]} is made from {[}T{]} & eiffel\_tower$_n^1$ & wrought\_iron$_n^1$ \\
 & P461 (Opposite of) & {[}H{]} is the opposite of {[}T{]} & technophilia$_n^1$ & technophobia$_n^1$ \\
 & P737 (Influenced by) & {[}H{]} is influenced by {[}T{]} & mozart$_n^1$ & bach$_n^1$ \\
 & P2283 (Uses) & {[}H{]} uses {[}T{]} & oil\_painting$_n^1$ & oil\_paint$_n^1$ \\
 & P463 (Member of) & {[}H{]} is a member of {[}T{]} & taiwan$_n^1$ & world\_trade\_organization$_n^1$ \\
 & P1535 (Used by) & {[}H{]} is used by {[}T{]} & rocket\_fuel$_n^1$ & rocket$_n^2$ \\
 & P279 (Subclass of) & {[}H{]} is a type of {[}T{]} & baroque$_n^1$ & expressive\_style$_n^1$ \\
\bottomrule
\end{tabular}
}
\caption{Templates used to verbalize triples from WordNet and WikiData. Not required for our ConceptNet subset.}
\label{tab:templates}
\end{table*}

Templates used for WordNet and WikiData triples are available in \autoref{tab:templates}.
In order to make predictions more consistent across sources, we found the most frequent determiners used with the head and tail terms of ConceptNet assertions and applied them on the WordNet and WikiData queries, wherever those same head and tail terms occurred.


\section{Extraction Statistics}
\label{sec:appendix_extraction}

\autoref{tab:ckg_stats} reports the relation counts for triples extracted using the procedure described on \autoref{sec:exp_extraction}.

\begin{table*}[t]
\centering
\begin{tabular}{lc}
\toprule
\textbf{Relation} & \multicolumn{1}{l}{\textbf{Count}} \\ \midrule
IsA & 6,557 \\
AtLocation & 5,104 \\
PartOf & 2,559 \\
UsedFor & 2,523 \\
MadeOf & 1,293 \\
Causes & 680 \\
CausesDesire & 663 \\
HasPrerequisite & 659 \\
HasA & 644 \\
CapableOf & 534 \\
MotivatedByGoal & 532 \\
HasProperty & 424 \\
CreatedBy & 390 \\
Desires & 229 \\
HasSubevent & 161 \\
ReceivesAction & 77 \\
HasLastSubevent & 53 \\
HasFirstSubevent & 6 \\
\bottomrule
\end{tabular}
\caption{Relation counts for novel triples extracted.}
\label{tab:ckg_stats}
\end{table*}


\section{SenseLAMA without gloss information}
\label{sec:appendix_senselama_no_gloss}

\autoref{tab:complete_results_no_gloss} reports results by relation on SenseLAMA (Full) when not prepending glosses or averaging sense embeddings with gloss embedding.
This is intended to better demonstrate which relations are most affected by the use of glosses, besides their overall impact on this probing task.

\begin{table*}[ht]
\centering
\resizebox{0.99\textwidth}{!}{%
\begin{tabular}{@{}l|ccccc||cccccc@{}}
\toprule
 & \multicolumn{5}{|c||}{\textbf{Core (4,960 candidates)}} & \multicolumn{6}{c}{\textbf{Full (117,659 candidates)}} \\
\multicolumn{1}{l|}{} & \multicolumn{1}{c}{\textbf{P@1}} & \multicolumn{1}{c}{\textbf{P@3}} & \multicolumn{1}{c}{\textbf{P@10}} & \multicolumn{1}{c}{\textbf{P@100}} & \multicolumn{1}{c||}{\textbf{MRR}} & \textbf{P@1} & \textbf{P@3} & \textbf{P@10} & \textbf{P@100} & \textbf{P@1000} & \textbf{MRR} \\ \midrule

\textbf{All} & 14.39 & 25.85 & 40.62 & 67.04 & 23.04 & 2.71 & 5.36 & 9.73 & 21.68 & 40.25 & 5.07 \\
 &  &  &  &  &  &  &  &  &  &  &  \\
 
\textbf{WordNet} & 17.53 & 29.88 & 36.48 & 44.39 & 26.46 & 3.00 & 5.69 & 9.89 & 20.86 & 37.83 & 5.32 \\
\textit{Hypernym} & 14.52 & 25.62 & 31.37 & 38.90 & 22.67 & 2.76 & 6.28 & 12.70 & 28.97 & 47.23 & 5.99 \\
\textit{Holonym (Member)} & 19.23 & 38.46 & 57.69 & 84.62 & 32.02 & 0.39 & 0.68 & 1.61 & 5.60 & 24.05 & 0.87 \\
\textit{Holonym (Part)} & 11.03 & 27.59 & 49.66 & 74.48 & 22.94 & 7.76 & 12.86 & 19.29 & 36.62 & 57.88 & 11.78 \\
\textit{Antonym} & 34.04 & 48.94 & 64.54 & 78.01 & 44.43 & 3.30 & 6.16 & 9.80 & 19.01 & 30.42 & 5.55 \\
\textit{Hypernym (Instance)} & - & - & - & - & - & 1.21 & 3.32 & 7.23 & 14.21 & 26.87 & 3.04 \\
\textit{Meronym (Substance)} & 25.00 & 43.75 & 62.50 & 87.50 & 37.89 & 0.91 & 1.67 & 3.80 & 16.87 & 38.15 & 2.11 \\
 &  &  &  &  &  &  &  &  &  &  &  \\
\textbf{WikiData} & 10.29 & 22.06 & 36.76 & 63.24 & 19.46 & 1.50 & 3.52 & 8.10 & 22.63 & 46.44 & 3.67 \\
\textit{P31 (Instance of)} & 5.13 & 10.26 & 23.08 & 46.15 & 10.51 & 0.67 & 2.26 & 5.19 & 16.81 & 37.30 & 2.31 \\
\textit{P361 (Part of)} & 11.11 & 20.00 & 44.44 & 73.33 & 20.85 & 1.98 & 4.02 & 9.44 & 27.21 & 54.06 & 4.43 \\
\textit{P366 (Use)} & 3.70 & 25.93 & 25.93 & 51.85 & 15.31 & 1.97 & 3.41 & 9.04 & 24.51 & 46.92 & 4.05 \\
\textit{P186 (Made from)} & 26.67 & 33.33 & 46.67 & 80.00 & 34.34 & 1.72 & 3.76 & 9.08 & 25.35 & 53.05 & 4.06 \\
\textit{P461 (Opposite of)} & 20.00 & 50.00 & 70.00 & 90.00 & 37.02 & 4.99 & 10.38 & 20.36 & 38.12 & 59.28 & 9.69 \\
\textit{P737 (Influenced by)} & - & - & - & - & - & 0.95 & 2.53 & 4.43 & 14.24 & 43.99 & 2.29 \\
\textit{P2283 (Uses)} & - & - & - & - & - & 1.12 & 2.99 & 8.21 & 20.15 & 44.78 & 3.31 \\
\textit{P463 (Member of)} & - & - & - & - & - & 1.64 & 6.01 & 14.75 & 46.99 & 84.15 & 5.91 \\
\textit{P1535 (Used by)} & - & - & - & - & - & 0.00 & 1.32 & 5.30 & 15.23 & 49.01 & 1.46 \\
\textit{P279 (Subclass of)} & - & - & - & - & - & 1.52 & 1.52 & 3.03 & 22.73 & 40.91 & 2.83 \\
 &  &  &  &  &  &  &  &  &  &  &  \\
\textbf{ConceptNet} & 9.59 & 19.48 & 34.69 & 67.64 & 17.67 & 1.84 & 5.20 & 11.21 & 29.23 & 55.78 & 4.93 \\
\textit{AtLocation} & 9.15 & 18.29 & 33.54 & 72.87 & 17.08 & 2.29 & 6.08 & 12.45 & 31.47 & 62.65 & 5.54 \\
\textit{UsedFor} & 4.32 & 12.96 & 29.01 & 66.05 & 11.99 & 1.51 & 3.63 & 9.53 & 27.08 & 56.43 & 4.09 \\
\textit{IsA} & 19.17 & 31.67 & 52.50 & 71.67 & 28.78 & 0.98 & 6.64 & 14.06 & 37.50 & 61.13 & 5.44 \\
\textit{Causes} & 7.89 & 13.16 & 21.05 & 39.47 & 12.51 & 0.89 & 1.79 & 4.91 & 16.96 & 35.27 & 2.22 \\
\textit{HasSubevent} & 0.00 & 8.77 & 17.54 & 29.82 & 5.88 & 0.98 & 0.98 & 2.44 & 10.24 & 21.46 & 1.49 \\
\textit{HasPrerequisite} & 8.00 & 16.00 & 32.00 & 62.00 & 14.71 & 1.21 & 4.24 & 10.30 & 29.09 & 53.94 & 4.37 \\
\textit{HasProperty} & 0.00 & 10.64 & 27.66 & 80.85 & 9.45 & 1.27 & 5.10 & 7.01 & 32.48 & 60.51 & 4.36 \\
\textit{CapableOf} & 6.25 & 12.50 & 22.92 & 66.67 & 12.29 & 0.81 & 3.25 & 11.38 & 21.95 & 47.97 & 3.82 \\
\textit{MotivatedByGoal} & 10.81 & 37.84 & 54.05 & 86.49 & 27.02 & 5.77 & 12.50 & 27.88 & 49.04 & 70.19 & 12.30 \\
\textit{HasA} & 14.58 & 16.67 & 39.58 & 62.50 & 20.76 & 3.09 & 10.31 & 14.43 & 31.96 & 62.89 & 7.28 \\
\textit{PartOf} & 39.39 & 51.52 & 63.64 & 93.94 & 47.47 & 11.25 & 16.25 & 30.00 & 42.50 & 75.00 & 16.43 \\
\textit{CausesDesire} & 0.00 & 0.00 & 14.29 & 50.00 & 3.41 & 0.00 & 0.00 & 1.92 & 13.46 & 30.77 & 0.63 \\
\textit{ReceivesAction} & 21.05 & 31.58 & 36.84 & 52.63 & 28.18 & 0.00 & 0.00 & 4.55 & 20.45 & 38.64 & 1.29 \\
\textit{MadeOf} & 5.56 & 38.89 & 55.56 & 88.89 & 24.40 & 0.00 & 0.00 & 9.52 & 33.33 & 57.14 & 2.63 \\
\textit{Desires} & 0.00 & 7.69 & 7.69 & 53.85 & 5.54 & 0.00 & 7.14 & 10.71 & 10.71 & 50.00 & 3.65 \\
\textit{CreatedBy} & - & - & - & - & - & 0.00 & 11.76 & 11.76 & 29.41 & 70.59 & 5.62 \\
\textit{HasFirstSubevent} & - & - & - & - & - & 0.00 & 0.00 & 0.00 & 50.00 & 85.71 & 2.27 \\
\textit{HasLastSubevent} & - & - & - & - & - & 0.00 & 0.00 & 0.00 & 16.67 & 41.67 & 0.85 \\
\bottomrule
\end{tabular}%
}
\caption{Complete results on the SenseLAMA probing task using BERT Large with LMMS SP-USM sense embeddings, not averaged with gloss embeddings and without prepending glosses to assertions, in contrast to \autoref{tab:complete_results}. Reporting Precision at k (P@k) and Mean Reciprocal Rank (MRR). Sorted by P@1 on the Full set.}
\label{tab:complete_results_no_gloss}
\end{table*}

\clearpage

\end{document}